# NOUVELLES DONNEES, NOUVELLES METHODES ? UN PANORAMA ET UNE ILLUSTRATION DE L'EMPLOI DE METHODES QUANTITATIVES INDUCTIVES POUR LA RECHERCHE EN GRH.

**Résumé :**

« *Data is the new oil* », autrement dit, les données seraient la source essentielle de la quatrième révolution industrielle en cours, ce qui a conduit certains observateurs à assimiler trop rapidement la quantité des données à une richesse en soi, et considérer l'avènement des big data comme une source quasi directe de profit. La GRH n'échappe pas à cette tendance, et l'accumulation de masses de données importantes sur les salariés apparaît dans le discours de certains entrepreneurs comme une condition nécessaire et suffisante pour la construction de modèles prédictifs de comportements complexes au travail comme l'absentéisme ou la performance en poste.

En réalité, l'analogie est un peu trompeuse : contrairement au pétrole, il n'y a pas ici d'enjeux majeurs concernant la production des données (dont les flux sont générés en continu et à faible coût par divers systèmes d'information), mais plutôt leur « raffinage » c'est-à-dire les opérations nécessaires à la transformation de ces données en produit utile, autrement dit en connaissance, si possible actionnable. C'est dans cette transformation que résident les enjeux méthodologiques de la valorisation des données à la fois pour les praticiens et pour les chercheurs académiques. Les réflexions sur les méthodes applicables pour tirer parti des possibilités offertes par ces données massives sont relativement récentes, et mettent souvent en avant l'aspect disruptif du déluge de données actuel pour souligner le fait que cette évolution serait à la source d'un renouveau de l'empirisme dans un « quatrième paradigme » fondé sur l'exploitation intensive et « agnostique » de masses de données en vue de faire émerger des connaissances nouvelles, selon une logique purement inductive. Sans adopter ce point de vue spéculatif, force est de constater que les approches inductives basées sur les données (*data driven*) demeurent rares dans les études quantitatives en GRH. Il existe pourtant des méthodes bien établies, notamment dans le domaine de la fouille de données *(data mining)*, qui reposent sur des approches inductives. Ce domaine de l'analyse quantitative à visée inductive demeure encore assez peu exploré en GRH (à l'exception des analyses typologiques). L'objectif de cette communication est tout d'abord de dresser un panorama des méthodes mobilisables pour des recherches en GRH, avant de proposer une illustration empirique qui consiste en une recherche exploratoire combinant une analyse en profils latents et une exploration par modèles graphiques gaussiens.

**Mots-clés** : Méthodologie quantitative, analyse exploratoire, induction, modèles graphiques gaussiens, analyse de profils latents

**Abstract :**

"Data is the new oil", in short, data would be the essential source of the ongoing fourth industrial revolution, which has led some commentators to assimilate too quickly the quantity of data to a source of wealth in itself, and consider the development of big data as an quasi direct cause of profit. Human resources management is not escaping this trend, and the accumulation of large amounts of data on employees is perceived by some entrepreneurs as a necessary and sufficient condition for the construction of predictive models of complex work behaviors such as absenteeism or job performance.

In fact, the analogy is somewhat misleading: unlike oil, there are no major issues here concerning the production of data (whose flows are generated continuously and at low cost by various information systems), but rather their "refining", i.e. the operations necessary to transform this data into a useful product, namely into knowledge. This transformation is where the methodological challenges of data valuation lie, both for practitioners and for academic researchers. Considerations on the methods applicable to take advantage of the possibilities offered by these massive data are relatively recent, and often highlight the disruptive aspect of the current "data deluge" to point out that this evolution would be the source of a revival of empiricism in a "fourth paradigm" based on the intensive and "agnostic" exploitation of massive amounts of data in order to bring out new knowledge, following a purely inductive logic. Although we do not adopt this speculative point of view, it is clear that data-driven approaches are scarce in quantitative HRM studies. However, there are well-established methods, particularly in the field of data mining, which are based on inductive approaches. This area of quantitative analysis with an inductive aim is still relatively unexplored in HRM ( apart from typological analyses). The objective of this paper is first to give an overview of data driven methods that can be used for HRM research, before proposing an empirical illustration which consists in an exploratory research combining a latent profile analysis and an exploration by Gaussian graphical models.

Nouvelles données, nouvelles méthodes ? un panorama et une illustration de l'emploi de méthodes quantitatives inductives pour la recherche en GRH.

# NOUVELLES DONNEES, NOUVELLES METHODES ? UN PANORAMA ET UNE ILLUSTRATION DE L'EMPLOI DE METHODES QUANTITATIVES INDUCTIVES POUR LA RECHERCHE EN GRH.

## Introduction :

Dans le domaine de l'analyse de données appliquée au management, l'émergence des big data associée aux avancées récentes des systèmes apprenants et de l'intelligence artificielle permet un développement accéléré de l'analytique RH, qui s'appuie sur des collectes de données plus nombreuses, plus variées et parfois moins structurées (comme par exemple l'activité des salariés sur les réseaux sociaux). Ce développement offre des perspectives nouvelles, à la fois pour les managers et les chercheurs, mais pose de nombreux défis (Faraj et al., 2018; George et al., 2014) . L'analytique RH est définie par Marler & Boudreau (2017, p. 15) comme un ensemble de pratiques RH rendues possible par les technologies de l'information, utilisant des analyses des données relatives aux processus RH, au capital humain, à la performance organisationnelle afin de permettre la prise de décision basée sur les données (*data driven decision making*). Parmi les défis à relever pour les promoteurs de l'analytique RH, certains concernent l'accès, l'utilisation et la protection des données (Angrave et al., 2016). D'autres, plus techniques, concernent les méthodes disponibles pour analyser ces données nouvelles (*big data analytics,* voir Sivarajah et al., 2017), ainsi que les enjeux épistémologiques associés à l'émergence d'une pratique scientifique guidée par les données. C'est ce volet méthodologique et épistémologique qui sera abordé dans la présente communication.

La période récente est marquée par un changement graduel dans la relation entre recherche et données. Selon (Kitchin, 2014) nous serions même en train d'assister à un changement de paradigme épistémologique, avec le passage d'un paradigme guidé par la connaissance (*knowledge driven*) vers un paradigme guidé par les données (*data driven*). Sans forcément souscrire à cette prophétie qui concerne prioritairement le cas des big data, force est de constater que la disponibilité et le volume des données collectées sur les individus s'accroissent. Des bases de données importantes résultant d'enquêtes transdisciplinaires (ex : European Social Survey, OFER-DARES sur le recrutement) sont aujourd'hui accessibles aux chercheurs en GRH et fournissent des informations bien adaptées pour des analyses exploratoires, car elles croisent en général de nombreuses variables et une grande quantité d'observations (Bhushan et al., 2019). De manière plus fondamentale, le mouvement de promotion des données ouvertes (ex : le portail data.gouv.fr en France ou le portail data.gov aux Etats-Unis), vise à permettre le libre accès à des données primaires au sein d'écosystèmes mêlant outils et intervenants publics et privés (Mabi, 2015). Sur le terrain, un nombre croissant d'entreprises utilisent des procédures qui conduisent au recueil de données sur les salariés ou les candidats, comme par exemple des enquêtes de climat social courtes à intervalles réguliers sur smartphones, la collecte automatique de CV en ligne, les entretiens de recrutement enregistrés, voire même les capteurs sensoriels portés par certains salariés (Levenson, 2018). L'analytique RH s'installe comme une branche bien identifiée au sein de la GRH, avec certains domaines d'application largement explorés, comme ceux liés au contrôle de gestion social *(HR scorecards)* et d'autres chantiers en développement rapide, comme le recrutement prédictif (Ajunwa et al., 2016; Oberst et al., 2020) ou les modèles d'apprentissage supervisé appliqués à l'étude des déterminants de la fidélisation (Atef et al., 2022).

Le chercheur en GRH se retrouve aujourd'hui dans une situation où il a des possibilités croissantes d'accéder à des données qui n'ont pas été collectées pour répondre à une question de recherche particulière, ce qui l'incite à envisager des approches et des modes de traitement adaptés à ces situations particulières où l'on trouve des données « en quête de théorie » pour reprendre l'expression d'Ullmann (1985). Les méthodologies habituelles déployées dans les études quantitatives en management sont souvent issues de domaines dans lesquels les travaux ont été menés selon une démarche hypothético-déductive sur des échantillons de taille plutôt modeste dans un design consistant à collecter une quantité limitée de données afin de tester un modèle pré-établi et mettre en évidence des effets d'influence statistiquement significatifs. Cette approche guidée par la théorie (*theory driven*) peut se révéler inappropriée lorsque la quantité d'observations et/ou de variables prises en compte devient importante, et plus encore lorsque le chercheur ne se trouve pas dans la situation où il s'appuie sur un modèle de recherche pré-établi (Maass et al., 2018). Il peut être alors intéressant d'adopter une autre perspective, très commune dans certaines sciences de la nature comme l'astronomie ou la météorologie, consistant à favoriser un raisonnement inductif en s'appuyant sur des méthodes quantitatives permettant de simplifier, ordonner et rendre intelligible des données multidimensionnelles[1].

Cette perspective n'est pas complètement nouvelle en management, et peut mobiliser une panoplie d'outils en plein développement. La fouille de données (*data mining*) s'appuie depuis longtemps sur des procédures qui permettent d'explorer des masses importantes de données dans une logique explicative et prédictive (Tufféry, 2012). Parmi les exemples de méthodes connues figurent les approches itératives de la régression permettant de sélectionner des modèles parcimonieux par élimination des variables non contributives, qui sont utilisées dans l'apprentissage supervisé (*machine learning*). On peut aussi mentionner les analyses typologiques qui permettent de faire émerger des groupes d'observations et des profils-type, à l'image des analyses en profils latents (Marsh et al., 2009). Le développement récent des analyses de réseau, et notamment des modèles graphiques gaussiens qui permettent de représenter des liens entre un jeu de variables continues en limitant les corrélations illusoires, offre de nouvelles perspectives aux analyses multivariées (Borsboom et al., 2021). Toutes ces méthodes partagent la caractéristique d'être multivariées et de reposer sur une logique inductive, guidée par les données (*data driven*).

La recherche quantitative en GRH s'est jusqu'alors peu appuyée sur cette perspective inductive, à l'exception notable des analyses « centrées sur les personnes » (*person centered*) qui reposent sur l'hypothèse fondamentale que les populations étudiées ne sont pas homogènes et recourent par principe aux analyses typologiques (Biétry & Creusier, 2017; Chou et al., 2015; Dai & De Meuse, 2013; Meyer et al., 2013). Or, il nous semble que l'évolution récente invite à élargir le spectre d'analyse, à la fois pour des raisons pratiques et méthodologiques. Sur le terrain, on constate que les pratiques des organisations tendent à utiliser de plus en plus de données pour alimenter des processus inductifs de décision, ce qui conduit certains auteurs à parler d'un mouvement vers un management *data-driven* (Lee et al., 2015; Miragliotta et al., 2018). Il y a de forts enjeux de recherche dans ce domaine où se creuse un fossé entre recherche et pratiques de management des ressources humaines (Woods et al., 2019). Sur le plan méthodologique, la diffusion de certaines méthodes compatibles avec les approches inductives demeure confidentielle dans le champs de la GRH francophone : on peut citer à titre d'exemple certains travaux mobilisant des modèles prédictifs par réseau de neurones pour analyser les prédicteurs de l'implication au travail ou les mouvements de personnel (Fantcho & Babei, 2016; Valéau & Trommsdorff, 2014).

---

[1] Précisons ici que par analyses quantitatives, nous n'entendons pas seulement analyse de données quantitatives : on peut parfaitement analyser des données qualitatives par des méthodes quantitatives, et il existe une panoplie de méthode quantitatives qui s'appliquent prioritairement à des données qualitatives (on peut citer l'analyse factorielle des correspondances ou l'analyse lexicale).

Nous proposons dans la présente communication de contribuer à une meilleure diffusion de ces méthodes, en présentant tout d'abord les principaux enjeux méthodologiques et épistémologique liés à l'approche inductive guidée par les données, puis en présentant un panorama synthétique des méthodes quantitatives particulièrement adaptées à une approche inductive guidée par les données en GRH. Nous illustrerons dans une seconde partie l'emploi combiné de trois types d'analyses (typologie, analyse prédictive et analyse de réseau) dans une recherche exploratoire sur une nouvelle mesure de l'implication au travail, qui a été menée sur un échantillon de 850 salariés.

## 1. Les approches quantitatives inductives en GRH

### 1.1. *Recherches theory driven et data driven : deux voies complémentaires*

Traditionnellement, la plupart des démarches scientifiques débutent par un questionnement : le chercheur s'appuie sur l'information disponible dans la littérature pour mettre en évidence un manque, une inconsistance ou un paradoxe dont la résolution permettra de faire avancer la connaissance. Deux voies de recherche sont alors possibles : explorer ou tester (Durieux & Charreire Petit, 2007). Il s'agit de deux démarches distinctes de collecte et d'analyse des données, dont les caractéristiques essentielles sont résumées dans le Tableau 1. L'approche *theory driven* suit une logique descendante, classiquement organisée sous forme hypothético déductive (Igalens & Roussel, 1998; Thiétart, 2007). L'approche *data driven* suit une logique opposée et ascendante, dans laquelle des données primaires ou secondaires constituent le socle sur lequel le chercheur va s'appuyer pour répondre à la question de recherche, ou même parfois s'engager dans une exploration sans a priori, démarche qualifiée « d'agnostique » par Jack et al. (2018). Son objectif est alors de rechercher des associations solides, des régularités, des groupes homogènes lui permettant de construire une interprétation par induction. Cette approche parfois présentée comme en rupture par rapport à la précédente (Andersen & Hepburn, 2019) s'est largement répandue dans les sciences de la nature (médecine, écologie), en profitant de la puissance de calcul des ordinateurs. Dans le champ spécifique de la GRH, les analyses typologiques (Morin et al., 2010; Strohmeier & Kabst, 2014; Wang & Hanges, 2011), les recherches qualitatives mobilisant la théorie ancrée (Galindo, 2017; Murphy et al., 2017), ou encore les approches QCA (*qualitative comparative analysis,* Delerue & Moisson, 2021; Stimec, 2018) constituent des exemples de ce type de démarche.

**Tableau 1** : Comparaison des approches inductives et déductives (adapté de auteur, 2021)

|  | Approche déductive *theory driven* | Approche inductive *data driven* |
|---|---|---|
| Objectif | -Expliquer<br>-Tester / Vérifier<br>-Généraliser (par inférence statistique) | -Explorer<br>-Rechercher des structures sous-jacentes (patterns), des régularités et des associations, rechercher la présence de sous-groupes et profils<br>-Prédire |
| Données | Données généralement primaires collectées en quantités limitée et dont la qualité est contrôlable | Données primaires ou secondaire, collectés en quantité limitée ou en grande quantité (*big data*), dont la qualité peut être variable |
| Processus de recherche | Top down (descendant)<br>Question de recherche issue de la littérature → Hypothèses → Design de l'étude → Collecte de données → Test → Inférence<br>(Andersen & Hepburn, 2015) | Bottom-up (ascendant)<br>Question de recherche → récupération de données primaires ou secondaires → nettoyage et préparation des données → recherche de structures (*patterns*) ou corrélations, identification de profils types (segmentation) → Interprétation des régularités ou des profils (Jagadish, 2015) |
| Critère d'évaluation | Validation ou rejet des hypothèses issues du modèle de recherche | -Capacité prédictive ou explicative du modèle<br>-Pertinence conceptuelle des profils obtenus (en cas d'analyses de segmentation) |
| Méthodes appliquées | -Méthode expérimentale<br>-Statistiques inférentielles<br>-Modèles de régression au sens large (ANOVA, régression, modèles linéaires généralisés, modélisation par équations structurelles) | -Analyses factorielles (ACP, ACM)<br>-Méthodes graphiques (positionnement multidimensionnel, analyse de réseau)<br>-Analyses typologiques (classification hiérarchique, centres mobiles, classes latentes)<br>-Apprentissage supervisé (*machine learning, deep learning*) |
| Avantages | -La standardisation des procédures permet une accumulation du savoir et une validation croisée des résultats (par exemple avec des méta-analyses). | -Possibilité de tirer parti de la puissance de calcul des outils d'analyse de données.<br>-Possibilité de traiter des populations (l'inférence devient dans ce cas inutile)<br>-Limitation des biais culturels et cognitifs du chercheur grâce à une posture « agnostique » face aux données (Jack et al., 2018) |
| Inconvénients | -Les cadres théoriques et hypothèses reflètent les biais culturels et cognitifs du chercheur (Holzleitner et al., 2019)<br>-Les cadres pré-établis limitent les approches transdisciplinaires (Maas & al, 2018)<br>-Les échantillon sont souvent faibles et coûteux à obtenir, ce qui peut nuire à la généralisation et la reproductibilité des résultats (Open Science Collaboration, 2015) | -Les résultats obtenus avec certaines méthodes sont parfois impossible à expliquer (phénomène de « boite noire » dans les algorithmes d'apprentissage profond ou non supervisé.<br>-Une démarche data driven « pure » non informée par un cadre théorique préalable est très délicate à envisager.<br>- La reproductibilité des résultats n'est pas garantie, notamment dans le cas de certaines méthodes non supervisées (ex : analyse typologique) dont les résultats comportent une part aléatoire liée au paramétrage et au fonctionnement des algorithmes |

## 1.2. L'induction fondée sur les données : « fin de la théorie dans les sciences » ou nouveau paradigme épistémologique ?

Les deux démarches présentées dans le point précédent ont progressé de manière parallèle. Malgré l'ancienneté reconnue de l'approche inductive à l'origine du courant empirique [2], certains auteurs insistent sur la rupture épistémologique fondamentale associée à l'émergence des approches *data driven* contemporaines appuyées sur l'accès à des masses de plus en plus importantes de données, qui conduirait à l'émergence d'un « nouvel empirisme technologique » (Arbia, 2021; Mazzocchi, 2015). D'autres mentionnent même la « fin de la théorie » dans les sciences (Anderson, 2008). Ce point de vue est largement critiquable et critiqué (voir par exemple Mazzocchi, 2015) mais il dénote bien une croyance dans le caractère disruptif de l'utilisation des big data dans la recherche scientifique. Kitchin (2014) suggère que l'époque actuelle marque le passage d'un paradigme *knowledge driven* vers un paradigme *data driven*, ce qui peut s'interpréter comme le passage d'un paradigme de la causalité à un paradigme de la corrélation (Cukier & Mayer-Schönberger, 2014) : ce « quatrième paradigme » serait fondé sur la génération d'hypothèses à partir de l'exploration intensive de données, au contraire du paradigme actuel fondé sur la génération ou l'utilisation de données pour tester des hypothèses. L'un des intérêts majeurs de cette démarche est qu'elle se révèle propice à la sérendipité, et peut ouvrir la possibilité de découvertes fortuite ou inattendues. La visualisation parait comme un élément important commun à la plupart des approches quantitatives inductives contemporaines. Loin d'être une activité secondaire, la visualisation est reconnue comme un pilier de la démarche scientifique, y compris dans le domaine du management (Bell & Davison, 2013; Few, 2021). Elle apparait comme un outil pédagogique et heuristique, considéré par Latour (1986) comme un moyen crucial au service du raisonnement et de la découverte des phénomènes naturels et sociaux.

Dans le champ de la GRH et du comportement organisationnel, les approches quantitatives *theory driven* demeurent pourtant à ce jour largement dominantes. Mis à part les travaux reposant sur des analyses typologiques, il est rare de trouver dans les revues de référence des recherches utilisant des méthodes quantitatives qui ne reposent pas sur un cadre théorique préalable solidement établi et la formulation d'une série d'hypothèses ou de propositions.

Il faut toutefois noter que l'induction et la déduction peuvent aussi être vues comme complémentaires (Maass et al., 2018), car leur articulation permet un processus itératif entre données et théorie et s'avère tout à fait compatible avec des designs de recherche de type abductifs dans lesquels le chercheur opère les allers-retours entre théorie et données (David, 1999; Shani et al., 2020; van Hoek et al., 2005). Cette boucle abductive est parfois présentée comme un moyen de réconcilier les approches inductives et déductives et de résoudre le « dilemme des savoirs » en sciences de gestion (Pellissier-Tanon, 2001, p. 56).

Le domaine de la GRH est concerné par cette évolution, car il est de plus en plus impacté par l'utilisation intensive de données comme support d'aide à la décision : l'analytique RH est un secteur en plein développement et les solutions permettent aujourd'hui d'aller plus loin que le simple reporting : on passe d'une vision descriptive à une vision explicative, voire prédictive (Coron, 2019; Marler & Boudreau, 2017). Si l'on s'intéresse au champ de la recherche académique, l'analytique RH présente une vraie proximité et une forte compatibilité avec le courant du management par la preuve (*evidence based management*) : les données sont le socle à partir duquel il est possible de sélectionner les meilleures pratiques et fournir ainsi une aide pratique et argumentative à la décision pour les managers (Coron, 2019; Olivas-Luján & Rousseau, 2010). Le paragraphe suivant est consacré à la présentation synthétique des méthodes particulièrement adaptées à ces approches inductives fondées sur les données[3].

---

[2] La célèbre phrase d'Isaac Newton « Hypotheses non fingo » (Je ne fais pas d'hypothèses) illustre le rôle prégnant des données empiriques et de l'induction dans la méthode scientifique depuis ses origines.

## 1.3. Les approches quantitatives inductives : un panorama des méthodes disponibles pour la GRH

### 1.3.1 Les approches centrées sur les variables et centrées sur les personnes

Pour des raisons pédagogiques, nous allons distinguer d'une part approches « centrées sur les variables » (*variable centered*), dans lesquelles l'objectif est l'étude des relations entre des variables (représentées par des scores moyens) sur un échantillon et d'autre part les approches « centrées sur les personnes » (*person centered*), dans lesquelles l'objectif est de mettre en évidence des groupes d'individus ou d'observations (profils-types ou classes) caractérisés par des structures de scores (patterns) relativement homogènes sur certaines variables. Cette distinction fait écho à la distinction entre approche inductive et déductive ; elle correspond aussi à une inflexion relativement récente des recherches en GRH et comportement organisationnel soutenue par des chercheurs de renom (voir par exemple Meyer et al., 2013; Morin et al., 2010). Ceux-ci s'appuient sur le postulat que dans de nombreux cas la population enquêtée est fondamentalement hétérogène, et que l'étude de cette hétérogénéité en mobilisant des analyses typologiques au sens large peut être riche d'enseignements. Dans le domaine de la GRH, une approche centrée sur les personnes présente un grand intérêt, car elle permet d'isoler des profils types (par exemple des profils d'implication au travail) que l'on peut associer à des politiques RH adaptées, voire personnalisées. Si l'on prend en exemple l'implication au travail, plusieurs travaux ont été menés autour de cette problématique, et ont permis de confirmer l'existence de profils type d'implication, se caractérisant par une combinaisons particulière des différentes composantes (Somers, 2010; Tsoumbris & Xenikou, 2010). On peut remarquer que les approches centrées sur les personnes apparaissent « naturellement » inductives, même si elles sont parfois utilisées dans un cadre déductif comme outil de test d'hypothèses (voir par exemple Alessandri et al., 2015). A l'inverse, les approches centrées sur les variables peuvent au premier abord apparaitre naturellement déductives, car elles reposent souvent sur le choix de « variables cibles » (variables indépendantes) que le chercheur cherche à expliquer, mais elles peuvent aussi être utilisées dans une logique inductive, notamment lorsqu'il s'agit d'explorer une combinaison de prédicteurs possibles pour une variable dépendante donnée : les modèles d'apprentissage machine (*machine learning*) ou d'apprentissage profond (*deep learning*) ont en général pour objectif d'isoler les variables les plus influentes permettant d'expliquer une variable cible, comme par exemple la rétention des employés (Marvin et al., 2021) ou le succès en formation (Jenkins et al., 2022).

### 1.3.2. Un cas particulier : l'analyse statistique de données textuelles

L'analyse statistique des données textuelles a aujourd'hui trouvé sa place dans tous les domaines des sciences de gestion. Sous ce vocable générique, on retrouve de nombreuses approches, généralement associées à des logiciels spécifiques : l'analyse lexicométrique (Lebart & Salem, 1994), l'analyse lexicale par contexte (Reinert, 2007), l'analyse automatique de contenu (Seignour, 2011), l'analyse de sentiment ou d'opinion (Boullier & Lohard, 2012). De manière générale, l'analyse statistique des données textuelles consiste à appliquer les méthodes statistiques habituelles (analyses de fréquence, classification, analyses factorielles, comparaisons intergroupes) à des unités textuelles.

---

[3] Il est important de préciser que, même si l'essentiel des travaux traitant du « renouveau » de l'approche inductive s'appuient sur l'exemple des big data, cette approche ne nécessite pas forcément la présence de données massives pour trouver des terrains d'application. Les données RH possèdent rarement les caractéristiques des big data en termes de volume, de variété ou de vélocité, ce qui n'empêche nullement de leur appliquer une approche inductive.

*1.3.3. Panorama synthétique des méthodes*

Le tableau 2 présente un panorama synthétique et non exhaustif des méthodes quantitatives inductives utilisables en GRH. Les principes généraux de chaque approche y sont succinctement présentés, avec le type de variables avec lesquelles elles sont utilisables, et leur cible principale d'utilisation (personne ou variables) et. Les outils logiciels permettant de mettre en application les différentes méthodes sont présentés en annexe 1. Notons que s'il existe des logiciels commerciaux largement utilisés et bien adaptés pour toutes les analyses précédemment mentionnées (parmi les plus utilisés, citons Mplus, SPSS, SAS ou Stata), nous avons privilégié les solutions logicielles libres, qui présentent le point commun de reposer sur la base logicielle R (R Core Team, 2021). Il s'agit soit d'applications dédiées (packages) destinées à être utilisées dans l'environnement R, soit de logiciels comportant une interface graphique et reposant également sur une base R (JASP, JAMOVI, Iramuteq). Ces derniers se révèlent faciles à utiliser et présentent donc une courbe d'apprentissage moins « pentue » que l'environnement R.

**Tableau 2**. Un panorama non exhaustif des méthodes quantitatives inductives pour la GRH

| MÉTHODE | DESCRIPTION | TYPE DE VARIABLE | CIBLE |
|---|---|---|---|
| **Corrélations :** Représentation des relations bivariées entre tous types de variables | | | |
| Matrices de corrélations | Différents coefficients de corrélation sont disponibles selon le statut des variables : r de Pearson (variables numériques), rho de Spearman et tau de Kendall (variables nominales). Lorsque les variables sont ordinales ou nominales, les corrélation tétrachoriques ou polychoriques les assimilent à des variables numériques ayant une distribution normale sous-jacente. | Tous les types de variables | Variable |
| Corrélogrammes | Représentation graphique de matrices de corrélation avec un code couleur permettant d'évaluer visuellement le sens et l'importance des corrélations | Tous les types de variables | Variable |
| Tableaux croisés et coefficients d'association | Représentation des relations bivariées entre variables nominales. Les coefficients Phi ou V de Cramer constituent un analogue au coefficient de corrélation pour les variables nominales | Nominales et Ordinales | Variable |
| **Régressions « exploratoires » et méthodes explicatives *data driven* :** Méthodes de sélection des prédicteurs les plus pertinents parmi un ensemble de prédicteurs pour une variable dépendante donnée (numérique, ordinale ou nominale) | | | |
| Régressions linéaires hiérarchiques (modèle linéaire général) | Méthode descendante, ascendante ou combinée (*stepwise*). Ces méthodes reposent sur l'usage d'algorithmes permettant de sélectionner de manière itérative les prédicteurs statistiquement significatifs pour une variable cible numérique ou nominale | Tous les types de variables | Variable |
| Analyse de dominance (importance relative) | Évaluation des variables les plus influentes (c'est-à-dire les plus contributives au R² du modèle) dans un jeu de prédicteurs | Tous les types de variables | Variable |

| Régressions pénalisées (Ridge, Lasso) | Méthode de régression permettant de pénaliser (supprimer) les prédicteurs les moins contributifs dans un jeu de données, afin de simplifier le modèle | Tous les types de variables | Variable |
|---|---|---|---|
| Modèles prédictifs adaptatifs de classification (pour variable dépendante nominale) et de régression (pour variable dépendante continue. | Méthodes de régression non linéaire (ex : réseaux de neurones, forêts aléatoires,…) dont l'objectif est d'optimiser par apprentissage machine la qualité de prédiction d'un modèle. Leur effet « boite noire » provient de l'usage d'algorithmes opaques, entraînant l'impossibilité d'accéder au processus exact de sélection des variables : les seuls résultats disponibles concernent l'influence relative des variables et la qualité prédictive du modèle | Tous les types de variables | Variable |
| Méthode Fuzzy set QCA | Recherche des combinaisons de variables explicatives les plus efficaces pour prédire une variable cible (généralement utilisée pour des variables qualitatives) | Nominales Ordinales | Variable |
| **Analyses factorielles :** Analyses non supervisées permettant de simplifier la représentation de jeux de données multidimensionnels par des méthodes de regroupement, accompagnées de représentations graphiques basées sur des projections dans un espace à 2 ou 3 dimensions (cartes factorielles) ||||
| Analyse en composantes principales (ACP) | A partir d'un jeu de données croisant des individus (en lignes) et des variables (en colonnes), l'ACP cherche à résumer l'information apportée par les variables par un nombre limité de composants principaux formés de variables corrélées | Numériques Ordinales | Mixte |
| Analyse factorielle des correspondances (AFC) | Représentation graphique d'un tableau croisé (tableau de contingence) permettant d'étudier les liens entre les modalités de deux variables qualitatives, avec la représentation des individus et des variables sur le même plan | Nominales | Mixte |
| AFCM Analyse factorielle des correspondances multiples | Extension de l'AFC au cas de plusieurs variables nominales (tableaux de contingence multidimensionnels) | Nominales | Mixte |
| Positionnement multidimensionnel (*Multidimensional scaling*) | Projection dans un espace à deux ou trois dimension dimensions d'une matrice de distance entre un nombre donné d'objets (par exemple des variables) | Tous les types de variables | Variables |
| **Analyses typologiques (clustering) :** Analyse non supervisée avec une logique analogue à celle des analyses factorielles, mais orientées sur les individus ou les observations (lignes). L'objectif est d'isoler un nombre fini de groupes d'observations homogènes à partir d'une matrice de distance entre les observations. ||||
| Classifications hiérarchiques | Méthode de classification reposant sur l'agrégation successive d'individus proches (selon un indicateur de similarité) | Numériques* | Personne |
| Classification par partitionnement : centres mobiles, *k-means, k-medioids…* | Méthodes de classification itératives dans lesquels les observations/individus sont agrégés selon leur degré de similarité autour d'un nombre initial d'individus choisis au hasard. L'objectif est de maximiser la qualité du regroupement pour un nombre de classes fixé à l'avance. | Numériques | Personne |
| Modèles de mélange (mixture modeling) : classes latentes et profils latents | Méthodes de classification reposant sur l'hypothèse que la population est composée de sous-groupes homogènes « mélangés ». Les algorithmes utilisés permettent d'isoler les sous-groupes (classes ou profils) et d'affecter les observations à des groupes en incorporant une part d'incertitude (probabilités d'appartenance) | Numériques | |
| Analyses typologiques longitudinales (analyses de | Extension de l'analyse typologique à des design longitudinaux. Trajectoires latentes : extension de l'analyse en classes latentes à des mesures longitudinales | Tous types de variables | Personne |

| séquences, analyse de trajectoires latentes) | Analyse de séquences : modélisation de séquences et recherche de profils basés sur des trajectoires-types (enchaînement de séquences) | | |
|---|---|---|---|
| **Réseaux psychologiques :** Outil de visualisation et d'analyse des interrelations entre variables. Possibilité d'utiliser des indicateurs de centralité et d'intermédiation permettant d'évaluer l'importance et le rôle des variables dans un réseau. | | | |
| Réseaux psychologiques | Réseaux constitués par une ensemble de variables (nœuds) connectées par des liens (corrélations, corrélations partielles) | Tous types de variables | Variable |
| Modèles graphiques gaussiens (MGG) | Réseaux psychologiques constitués à partir de corrélations partielles entre variables continues (les corrélations partielles permettent de raisonner sur des corrélations « toutes choses égales par ailleurs », c'est-à-dire de limiter les cas de corrélations illusoires causées par des variables de confusion. | Numériques | Variable |
| Modèles Ising | Réseaux psychologiques constitués à partir de variables binaires | Nominales (binaires) | Variable |
| **Statistiques textuelles :** Ensemble de méthodes statistiques appliquées à des données textuelles (nominales). L'unité d'analyse peut être le mot (lexicométrie), ou des unités de sens comportant plusieurs mots | | | |
| Analyses lexicométriques | Utilisation d'outils analytiques (calculs de fréquences, analyses de spécificités, classification hiérarchique) ou visuels (Nuages de mots, cartes factorielles, graphes de co-occurrence) afin de caractériser un corpus textuel. Le corpus peut être un ensemble de textes ou un ensemble de mots, accompagnés de descripteurs (variables catégorielles permettant de réaliser des statistiques de comparaison par groupes). | Nominales (texte) | Mixte |
| Classification descendante (méthode Reinert) | Méthode de classification d'unités de contexte (mots ou ensembles de mots) permettant de faire émerger des groupes de discours caractérisés par des similitudes de vocabulaire | Nominales (texte) | Mixte |
| Analyse de sentiment | Analyse des sentiments associés aux lexique (ce type d'analyse nécessite de disposer de dictionnaires ou lexiques de sentiments) | Nominales (texte) | Personne |
| **Segmentation prédictive** | | | |
| *Predicted oriented segmentation* | Analyse typologique basée sur des modèles de régression PLS (recherche de sous-groupes se différenciant au niveau des coefficients des modèles structurels). Plusieurs approches disponibles : FIMIX, PLS-REBUS | Numériques | Mixte |

*\*Les algorithmes de classification opèrent sur des matrices de distance, qui sont toujours numériques. Ceci posé, il existe des possibilités de calculer des distances entre variables nominales en utilisant des métriques appropriées, ce qui permet in fine de réaliser des classifications sur tous les types de variables.*

Nous allons dans la partie suivante proposer un exemple de l'application conjointe de plusieurs des méthodes citées dans l'exploration d'un jeu de données collectées après d'un échantillon de 850 salariés dans le but de développer une nouvelle mesure de l'implication au travail.

## 2. Illustration empirique : exploration de profils d'implication
### *2.1. Contexte de l'étude et échantillon*
La base utilisée pour notre illustration empirique provient d'une étude réalisée dans le cadre d'un projet de validation d'une nouvelle mesure de l'implication au travail proposée par Klein et al. (2014). L'argument fondamental avancé par Klein et ses collègues pour proposer de redéfinir le concept d'implication au travail est que celui-ci est défini de manière trop floue, que ses échelles de mesure ne sont pas suffisamment centrées sur la notion précise d'implication, et qu'elles évaluent parfois des concepts voisins, comme l'identification organisationnelle ou la volonté de rester dans l'organisation. La nouvelle définition proposée conceptualise l'implication comme « un lien psychologique volontaire qui reflète le dévouement et la responsabilité envers une cible particulière » (Klein et al., 2012, p. 137). Elle s'accompagne d'une échelle de mesure resserrée, comportant 4 indicateurs et susceptible d'être appliquée à toutes les cibles habituelles de l'implication (voir Annexe 2). Cette échelle a été traduite en français et a été adaptée à trois cibles traditionnellement utilisées dans les études sur l'implication : les collègues de travail (cible proximale), l'organisation (cible médiane) et la profession (cible distale). Elle a ensuite fait l'objet d'une validation en utilisant les méthodes classiques recommandées dans la littérature (Roussel, 2005) : analyse factorielle exploratoire (factorisation par maximum de vraisemblance avec rotation promax), puis analyse confirmatoire par méthodes d'équations structurelles, suivie d'une évaluation de l'homogénéité, de la validité convergente et discriminante des trois facettes, et conclue par une analyse de la validité prédictive. Cette dernière a été établie en mettant en relation les trois échelles d'implication avec des construits psychologiques proches mais conceptuellement distincts : la satisfaction au travail, l'engagement dans le travail, l'identification organisationnelle et l'intention de départ.

## 2.2. Échantillon, mesures et estimation initiale de validité

L'échantillon de l'étude est composé de 850 salariés francophones (tableau 3) interrogés dans quatre pays (France, Belgique, Suisse et Canada), par un institut d'études en ligne.

**Tableau 3.** Caractéristiques de l'échantillon

| | |
|---|---|
| Sexe | H : 43.8 % ; F : 56.2 % |
| Age | Moyenne : 40.8 ; Ecart-type : 10.9 ; min : 19 ; max : 63 |
| Ancienneté | Moins de 2 ans : 23.3 % ; plus de 2 ans : 76.7 % |
| Position | Manager : 21.4 % ; non manager : 78.6 % |
| Taille de l'entreprise (nombre de salariés) | 0-9 : 17.9 % ; 10-49 : 20.3 % 50-199 : 20.2 % ; 200-999 : 20.3 % ; plus de 1000 : 21.3% |
| Pays (répartition) | France : 25.5% ; Canada : 25.8% ; Belgique : 23.6% ; Suisse : 25% |

Outre les trois échelles mentionnées précédemment (KUTO : implication organisationnelle, KUTC : implication envers les collègues et KUTPR : implication dans la profession), nous avons utilisé des échelles précédemment validées pour construire un réseau nomologique permettant l'étude de validité prédictive : la satisfaction générale au travail (Cammann et al., 1983), l'intention de quitter l'organisation (Dwivedi, 2015), l'identification organisationnelle (Mael & Ashforth, 1992), et l'engagement dans le travail (Schaufeli et al., 2006)..

Les résultats de l'analyse exploratoire et confirmatoire montrent que toutes les échelles sont unidimensionnelles, avec des indices d'homogénéité élevés, une variance moyenne extraite supérieure au seuil recommandé de 50%, et des coefficients factoriels (loadings) supérieurs à 0.7 (voir annexe 2). Les indicateurs d'ajustement du modèle confirmatoire global contenant les 8 échelles sont bons, si l'on se réfère aux seuils habituellement recommandés (Gana & Broc, 2018; Kline, 2011) : Khi² (390) = 927, $p < .001$ ; CFI = 0.96 ; TLI = 0.95 ; RMSEA = 0.05 ; IC RMSEA90% [0.042, 0.050] ; SRMR = 0.03. Pour conclure cette phase de validation, nous avons créé des variables composites en calculant pour chaque échelle de mesure, les scores prédits par le modèle (scores factoriels confirmés) qui seront utilisés comme variables de départ dans la suite des traitements.

## *2.3. Méthodologie d'analyse inductive combinée*

Nous adoptons ici une perspective inductive en trois temps, avec une première étape centrée sur les personnes visant à isoler des profils types, suivie par deux étapes centrées sur les variables comprenant la description et l'étude des prédicteurs d'appartenance aux profils puis l'analyse des interrelations entre les variables caractéristiques des profils identifiés.

Bien que nous ayons adopté un cadre général inductif, il est indispensable de mobiliser un modèle général de référence lorsque l'on envisage une approche centrée sur les personnes, car la classification et la recherche de profils-types sera effectuée à partir d'une série de variables dont le choix est crucial (Bauer, 2021). Notre point de départ est le même que celui adopté dans plusieurs études sur l'implication centrées sur les personnes : il s'agit d'isoler des profils d'implication et d'étudier leur compatibilité. La mesure dominante de l'implication, le modèle tri-dimensionnel d'Allen & Meyer(Allen & Meyer, 1996), a été plusieurs fois investigué (Cooper et al., 2016; Meyer et al., 2013), ce qui a permis de mettre en évidence des profils-types caractérisés par des combinaisons spécifiques des dimensions affectives, continue et normative. En cohérence avec cette approche, notre objectif est ici d'identifier et analyser des profils-types fondés sur des configurations spécifiques des trois cibles d'implication[4].

Nous débutons par une analyse typologique en profils latents, avec pour objectif de faire émerger des profils caractérisés par une combinaison spécifique des trois cibles d'implication. Dans un deuxième temps, nous procédons à une exploration détaillée des profils identifiées en lien avec une série de variables illustratives (n'ayant pas servi pour la classification). Enfin, pour chaque profil, nous construisons un réseau des interrelations (modèle graphique gaussien) pour comparer les structures des relations entre les variables. L'objectif est d'accumuler des informations issues de différentes méthodes afin de parvenir à identifier des caractéristiques saillantes des profils d'implication. Les analyses ont été menées dans l'environnement R avec les packages mclust, tidy LPA, bootnet, relaimpo et FactoMiner (cf annexe 1 pour les références). Pour ne pas surcharger la présente communication, la présentation des méthodes utilisées et les détails de calculs figurent dans les annexes 2 à 6.

### *2.3.1 Analyse en profil latents*

L'analyse en profils latents est une forme de classification basée sur des modèles de mélange, qui présente deux avantages par rapports aux méthodes habituelles (classification hiérarchique ou modèles de partitionnement). Le fait que l'analyse repose sur une modélisation en variables latentes permet d'utiliser des critères d'ajustement des modèles aux données, qui peuvent servir d'aide au choix du nombre de classes. Le second avantage de cette méthode réside dans la possibilité d'incorporer un degré d'incertitude dans la classification par le biais des probabilités d'appartenance aux classes. Le terme d'analyse en classes latentes est réservé au cas de variables de classifications catégorielles et l'analyse en profil latents concerne les variables continues (Bauer, 2021).

La première décision importante est le choix des variables sur lesquelles va porter la classification : ce choix doit refléter les objectifs du chercheur : nous choisissons donc ici de construire les profils à partir des trois cibles d'implication.

---

[4] La mesure de Klein & al est posée comme unidimensionnelle. L'exploration des profils d'implication ne peut donc se faire qu'à un niveau supérieur, celui des cibles d'implication.

Les scores factoriels confirmés sauvegardés dans l'étape de validation des échelles sont utilisés pour lancer la classification. Plusieurs modèles ont été testés séquentiellement, puis ont été comparés sur une batterie d'indicateurs d'ajustement (voir annexe 3 pour les détails). En se basant sur les indicateurs d'ajustement et l'équilibre des profils, nous sélectionnons un modèle avec 5 profils, qui est bien équilibrée en matière d'effectifs de profils (graphique 1).

**Graphique 1.** Profils latents élaborés à partir de trois cibles d'implication

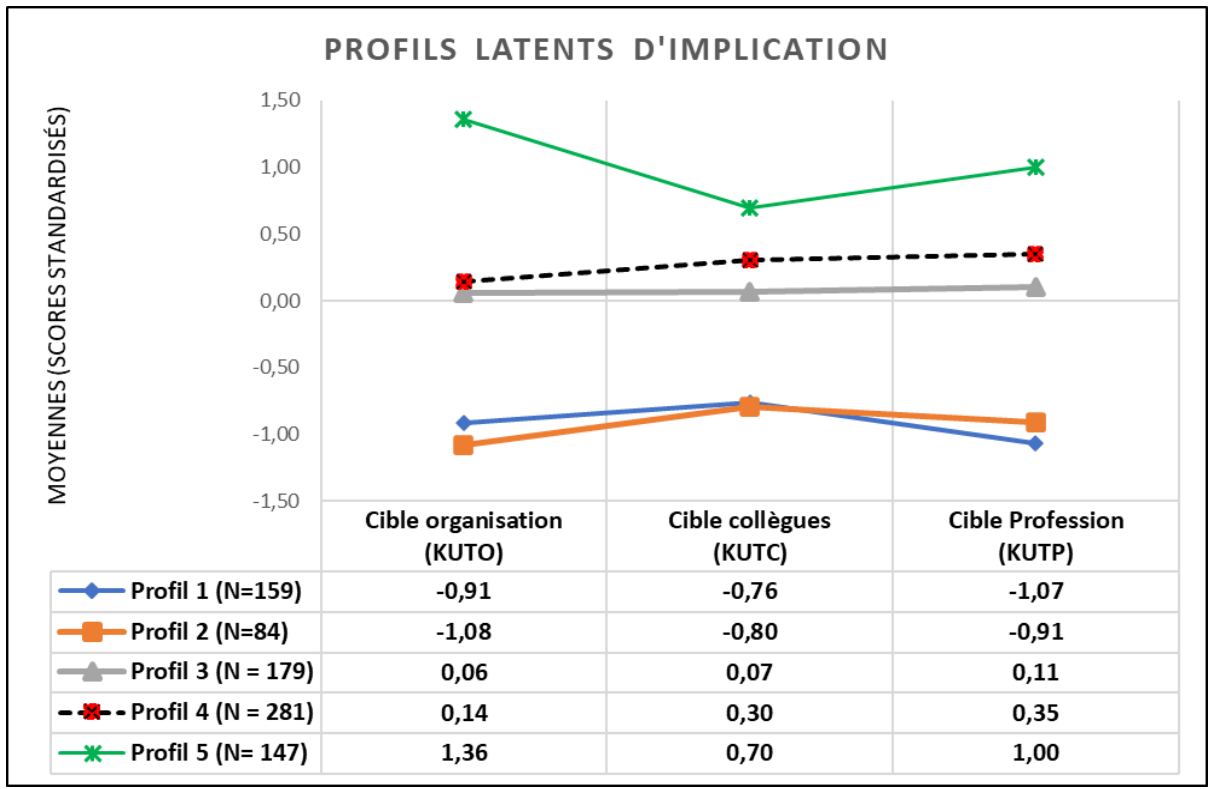

On remarque que certains profils sont très proches (1-2, 3-4), et que les profils ne se croisent pratiquement pas. Cette dernière caractéristique suggère une propriété particulière du construit mesuré (voir Gillet et al., 2021) : les profils ne se distingueraient pas selon leurs structures mais selon leur niveau, car on distingue une implication faible, moyenne et forte envers les trois cibles. Ceci suggère que les trois cibles reflètent peut-être un construit latent d'implication au sens large qui se traduirait par une propension générale à s'impliquer quelle que soit la cible. La présence d'un construit d'implication global est en cohérence avec les objectifs de Klein & al (2014) qui proposent un construit d'implication unidimensionnel et indépendant des cibles (*KUT = Klein & al Unidimensional Target free measure of commitment*). Une exploration subséquente du modèle d'implication multiple par le biais de modèles bi-factoriels pourrait être intéressante : elle permettrait d'utiliser à la fois un construit global et ses sous-dimensions dans le même design (voir par exemple Perreira et al., 2018)

*2.3.2. Exploration des profils identifiés*
Il est possible d'adopter une approche purement descriptive en analysant les différences entre profils sur des variables illustratives qui n'ont pas été utilisées pour la classification. Le chercheur peut réaliser une série d'ANOVA en utilisant la variable profil comme facteur et les variables illustratives comme variables dépendantes. Nous utilisons ici une approche un peu différente (implémentée dans le package FactoMiner) dans laquelle les profils sont caractérisés à l'aide de variables illustratives quantitatives et qualitatives : l'objectif est d'isoler les variables dont le score ou la fréquence sont statistiquement différentes sur le profil *par rapport à l'ensemble de l'échantillon* (Husson, Lê, et al., 2016). Cela permet d'isoler les modalités surreprésentées ou sous-représentées pour les variables qualitatives et de repérer les variables quantitatives dont les scores sont significativement supérieurs ou inférieurs à la moyenne générale dans un profil donné (voir détails en annexe 4).

**Tableau 4.** Description des profils par des variables illustratives

|  | Profil 1 | Profil 2 | Profil 3 | Profil 4 | Profil 5 |
|---|---|---|---|---|---|
| *Variables quantitatives* | | | | | |
| Turnover | + | + | | | - |
| Engagement | - | - | | + | + |
| Satisfaction | - | - | + | + | + |
| Identification | - | - | | + | + |
| Age | - | | + | | |
| *Variables qualitatives* | | | | | |
| Position hiérarchique | | | | Manager : + | |
| Pays | France : -<br>Canada : + | France : -<br>Canada : +<br>Suisse : + | | France : + | France : +<br>Canada : - |
| Ancienneté | < 1an : +<br>5 à 10 ans : - | | | | < 1 an : - |
| Taille de l'entreprise | | | | | >1000 salariés : - |

*Légende : le signe + signale une moyenne de score significativement supérieure sur le profil par rapport à la moyenne générale de l'échantillon ou une modalité sur-représentée, le signe – signale une moyenne significativement plus faible ou une modalité sous-représentée. Pour les cases grisées, le score ou la répartition ne se différencient pas de la moyenne générale.*

Les premiers examens montrent une relative homogénéité des profils deux à deux, en cohérence avec l'analyse précédente : les profils 1 et 2 décrivaient des salariés peu impliqués ; on voit ici qu'ils sont aussi insatisfaits, moins engagés et plus désireux de quitter l'organisation. L'examen des variables qualitative permet par exemple de constater que les salariés canadiens et les salariés débutants sont surreprésentés dans les profils « désimpliqués » et les salariés français sont surreprésentés dans les profils « impliqués ».

Dans le contexte d'une analyse en profil latents cette approche descriptive est parfois critiquée car elle ne permet pas de tirer parti de l'incertitude associée aux classes (Bakk & Kuha, 2021). On complètera donc ici la description avec une analyse exploratoire par une série de régressions linéaires visant à identifier les prédicteurs d'appartenance aux profils (en utilisant les probabilités d'appartenance aux différents profils comme variables dépendantes). Nous utiliserons une procédure exploratoire permettant d'identifier les prédicteurs les plus contributifs qui est appelée analyse de dominance ou d'importance relative[5]. Les analyses d'importance relative et de dominance ont le même objectif : il s'agit d'isoler au sein d'un groupe de prédicteurs potentiels ceux dont la puissance explicative, mesurée par leur contribution au $R^2$ du modèle (dans le cas d'une régression linéaire) est la plus importante.

---

[5] Nous aurons pu ici utiliser des approches de régression non linéaires par apprentissage machine (machine learning) qui ont aussi pour but d'isoler les variables influentes dans un jeu de prédicteurs.

Lorsque les prédicteurs ne sont pas corrélés, le calcul est trivial (le R² du modèle complet est la somme des R² de chaque modèle univarié). La situation est plus complexe lorsque les prédicteurs sont corrélés : les analyses d'importance relatives permettent dans ce contexte de calculer l'impact de chaque régresseur sur le R² en tenant compte de ses éventuelles interactions avec les autres prédicteurs (Grömping, 2006; Johnson & LeBreton, 2004). On doit ici noter que l'importance est donnée en pourcentage de contribution au R², ce qui n'informe pas sur le signe du prédicteur. Il est nécessaire d'examiner les corrélations ou construire un modèle de régression pour pouvoir s'assurer de la direction de l'influence.

Le tableau 5 résume les résultats obtenus pour les profils sur tous les prédicteurs du jeu de données. On remarque que les variables d'implication apparaissent naturellement comme les plus influentes, mais que le profil 3 est très mal expliqués par les prédicteurs (le R² est très faible). Ce mauvais ajustement signale que ce profil pourrait être artificiel (*spurious*), dans le sens où il résulterait par exemple de l'agrégation d'individus non classés dans les autres profils. La distinction entre « vrais » profils et profils artificiels est une question délicate dans le domaine de l'analyse en profils latents (Spurk et al., 2020). Dans une telle situation, il convient d'être prudent dans l'interprétation du profil suspect, ou tenter de répliquer la structure identifiée sur un autre échantillon. Dans le cas présent, nous choisissons de ne pas tenir compte de ce profil dans les calculs futurs car sa prédictibilité est très faible.

**Tableau 5.** Analyse de l'importance relative des prédicteurs (en % du R²)

| Prédicteurs | Prob prof. 1 R² : 0,49 | Prob prof. 2 R² : 0,17 | Prob prof. 3 R² : 0,02 | Prob prof. 4 R² : 0,14 | Prob prof. 5 R² : 0,42 | Influence moyenne |
|---|---|---|---|---|---|---|
| Age | 0,7% | 0,1% | 20,4% | 0,4% | 0,2% | **4,3%** |
| Pays | 0,8% | 7,8% | 20,2% | 4,3% | 0,8% | **6,8%** |
| Taille de l'entreprise | 0,6% | 3,2% | 11,7% | 2,9% | 1,1% | **3,9%** |
| Engagement | 8,8% | 6,3% | 1,5% | 2,2% | 11,8% | **6,1%** |
| Identification | 7,5% | 2,4% | 1,5% | 2,4% | 5,4% | **3,9%** |
| Implication collègues | 13,5% | 14,7% | 1,2% | 24,7% | 5,3% | **11,9%** |
| Implication organisation | 19,4% | 37,9% | 1,1% | 4,1% | 49,1% | **22,3%** |
| Implication profession | 36,0% | 17,7% | 5,0% | 41,0% | 13,7% | **22,7%** |
| Position hiérarchique | 0,1% | 0,2% | 14,1% | 7,4% | 0,6% | **4,5%** |
| Satisfaction | 8,2% | 3,4% | 12,9% | 2,6% | 5,5% | **6,5%** |
| Ancienneté | 1,9% | 4,8% | 8,4% | 2,5% | 1,1% | **3,7%** |
| Intention de quitter | 2,4% | 1,4% | 1,9% | 5,5% | 5,3% | **3,3%** |

*Légende : en grisé, les prédicteurs les plus influents pour chaque profil*

Nous remarquons que les profils qui apparaissaient très proches sur le graphique 1 (profils 1 et 2) n'ont pas les mêmes prédicteurs dominants : le niveau d'implication dans la profession est davantage prédicteur de l'appartenance au profil 1 et le niveau d'implication dans l'organisation est davantage prédicteur de l'appartenance au profil 2. L'examen des corrélations entre variables et profils (annexe 5) montre que ces deux prédicteurs sont négativement corrélés aux profils 1 et 2 et positivement liés aux profils 4 et 5 : une hausse de l'implication professionnelle réduit donc par exemple la probabilité d'appartenir au profil 1 et augmente la probabilité d'appartenance au profil 4.

Une fois les profils décrits par les variables illustratives, nous allons à présent nous intéresser aux liens entre les variables pour chacun des profils retenus.

*2.3.3. Visualisation des réseaux de relation entre variables*

L'approche la plus simple consisterait à construire des matrices de corrélation pour chaque profil et à comparer les corrélations entre les profils. Le problème est que ces matrices sont difficiles à interpréter lorsque les variables sont nombreuses, et que certaines corrélations peuvent être artificielles (*spurious corrélations*), c'est-à-dire causées par une variable de confusion. Nous allons utiliser ici une approche novatrice en GRH qui permet à la fois de visualiser les liens entre variables et de contrôler les corrélations factices. Il s'agit des modèles graphiques gaussiens (MGG), qui représentent les liens entre variables sous la forme d'un réseau dans lequel les nœuds (variables gaussiennes) sont reliés par des liens qui représentent des corrélations partielles. L'intérêt majeur des MGG est qu'ils reposent sur ces corrélations partielles : ceci signifie que la force du lien entre deux variables est établi *après avoir contrôlé l'effet de toutes les autres variables du réseau*. Cette propriété est intéressante car elle réduit le risque de mettre en évidence des corrélations factices (Bhushan et al., 2019, p. 2). L'absence le lien entre deux variables A et B apparait alors très informatif, puisqu'il signifie qu'A et B n'entretiennent aucune relation, une fois que l'on a pris en compte toutes les variables du réseau susceptibles de jouer le rôle de cause commune : on dit dans ce cas qu'A et B sont conditionnellement indépendants.

La représentation graphique d'un MGG incorpore deux informations importantes : la force de la relation, matérialisée par l'épaisseur du lien et le signe de la relation, matérialisé par une couleur (généralement le bleu ou le vert pour les liens positifs et le rouge pour les négatifs). L'intérêt majeur de la visualisation est de mettre en évidence des relations peu explorées jusqu'alors, voire surprenantes ou inattendues (Bhushan et al., 2019), mais également de mettre en évidence l'absence de relations directes entre construits (indépendance conditionnelle). L'usage de procédures de régularisation (suppression des faibles coefficients) permet si besoin de simplifier la représentation visuelle pour isoler les relations les plus importantes entre variables. L'usage d'indicateurs de centralité issus de l'analyse des graphes permet également d'apporter des informations supplémentaires sur le rôle et le poids des variables dans le réseau (pour une présentation pédagogique et complète, on peut se référer à Borsboom et al., 2021)

Dans le cas présent, nous allons utiliser les MGG pour comparer la structure des relations entre variables dans les différents profils. Les détails des indicateurs et les MGG estimés pour chaque profil figurent en annexe 6. L'intérêt des MGG dans le cadre de notre étude est exploratoire : on peut comparer des profils éloignés (ex : 1 et 5) ou au contraire étudier des profils apparemment proches (ex : 1 et 2) pour mettre en évidence des différences qui n'apparaissent pas clairement jusqu'alors[6]. A titre d'exemple nous allons analyser les MGG des profils 1 et 2, qui sont peu impliqués de manière globale, et assez proches dans toutes les analyses réalisées jusqu'alors. On constate pourtant visuellement que les réseaux estimés sur les deux profils sont très différents (graphique 2).

---

[6] La significativité des différence de magnitude entre les liens peut être estimé par une procédure de rééchantilonnage (voir (Epskamp et al., 2018)

**Graphique 2**. Modèles graphiques gaussiens estimés sur les profils 1 et 2 (Régularisation par la procédure EBIC Glasso)

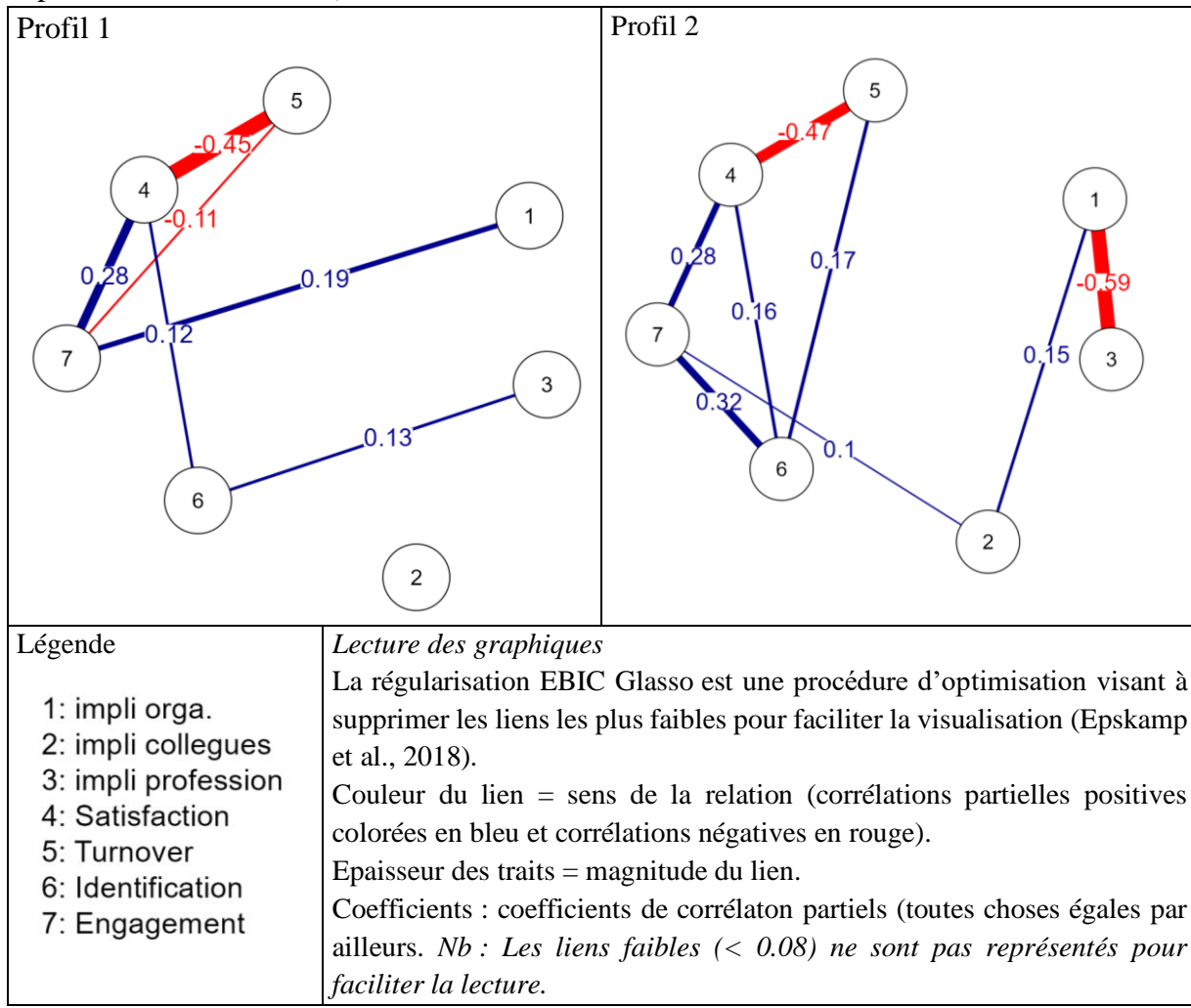

Les trois cibles d'implication semblent indépendantes dans le profil 1 (on ne trouve aucun lien direct entre les elles) et apparaissent au contraire substituables dans le profil 2 : les répondants semblent choisir entre implication organisationnelle (et envers les collègues dans une moindre mesure) et professionnelle, car il apparait forte corrélation négative entre les 2 cibles. On note aussi que l'implication envers les collègues joue un rôle plus important pour le profil 2 : elle semble jouer un rôle médiateur entre l'implication organisationnelle et l'engagement. Identification et engagement sont liés dans le profil 2, et on remarque que l'engagement agit directement sur l'intention de départ dans le profil 1. On note enfin un lien positif paradoxal entre identification et turnover pour le profil 2, mais on peut remarquer que le niveau d'identification est faible dans ce profil. Un lien positif entre identification et turnover peut être comme une forme de résignation et d'indifférence au sens de Klein & al (2012, p. 134) : les salariés ne s'identifient pas à l'organisation mais ne souhaitent pas pour autant la quitter. Cette attitude correspond à une forme d'implication de continuité dans le modèle d'Allen & Meyer (1996).

Si on s'intéresse au rôle des variables dans les deux réseaux (graphique 3), les indicateurs de centralité permettent de constater l'importance de la satisfaction dans les deux réseaux (fort degré de centralité, correspondant à de forts liens avec les construits voisins dans le réseau). Le degré de centralité des variables est proche pour les profils 1 et 2, mais on note que l'implication organisationnelle joue un rôle plus important dans le réseau associé au profil 2, et que l'engagement joue un rôle plus important dans le réseau du profil 1.

Les deux profils se distinguent davantage sur la centralité d'intermédiation (la capacité d'une variable à jouer le rôle de passerelle entre les différentes parties du réseau): la satisfaction joue un rôle de passerelle plus important pour le profil 1 ; le même rôle est joué par l'engagement dans le profil 2 (on remarque par exemple qu'il joue un rôle de passerelle entre l'implication envers les collègues et la satisfaction).

**Graphique 3.** Indicateurs de centralité

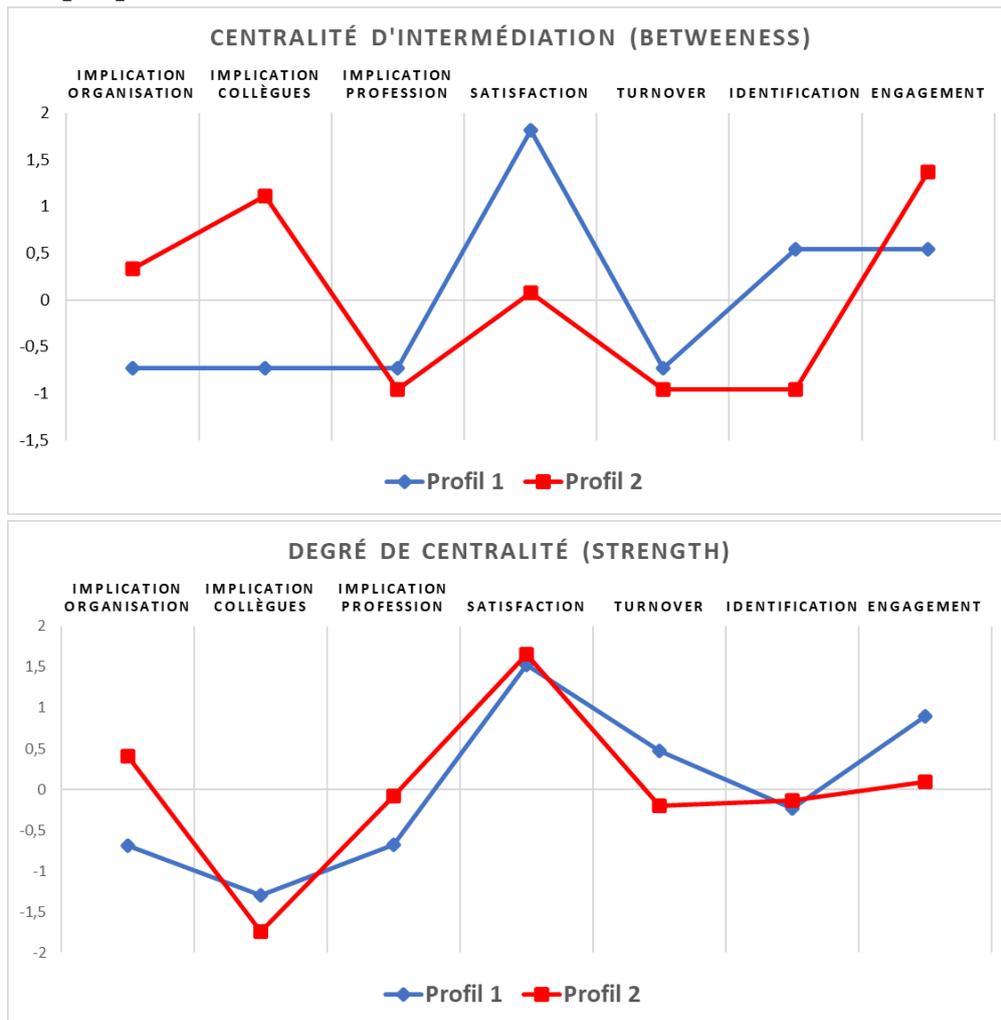

## 2.4. Synthèse et discussion

Le tableau figurant en annexe 7 présente une brève synthèse de l'analyse exploratoire menée. Notre discussion portera sur les aspects épistémologiques, techniques et managériaux associés à l'emploi d'un cadre inductif *data driven*.

### 2.4.1. L'approche inductive : des enjeux épistémologiques

L'induction est une démarche difficile, qui ne saurait se résumer à « faire apparaître » des lois générales : Pellissier Tanon (2001) remarque avec justesse que le raisonnement inductif repose sur une sélection et une organisation rigoureuse des faits observés et que « l'art du chercheur » consiste à appréhender les ressemblances et dissemblances pour en tirer des conjectures (ibid, p.9). En résumé, explorer et prédire n'est pas comprendre : les risques d'un empirisme technologique « aveugle » basé sur les données est que le déploiement des outils et l'application pratique précèdent la théorie et la réflexion. L'exemple des outils de sélection du personnel basés sur les algorithmes apprenants est typique de ce mouvement : certaines promesses d'inférences de compétences issues de la détection des expressions faciales en entretien vidéo sont très peu fondées scientifiquement et sans doute vecteur de discrimination (Amadieu, 2019). La « fin de la théorie » prophétisée par Anderson (2018) n'est sans doute pas encore d'actualité. En pratique, envisager une approche inductive « pure », uniquement basée sur les données comme cela est parfois suggéré est presque impossible : l'absence de cadre théorique préalable pose par exemple des problèmes pratiques pour toutes les méthodes dans lesquelles il faut choisir des variables pour débuter une exploration (comme c'est le cas avec l'analyse en profils latents utilisée dans la présente communication).

*2.4.2. S'appuyer sur les données : des enjeux techniques souvent éludés*
Les travaux classiques d'A. Desrosières (Desrosières, 2013; Desrosières & Kott, 2005), à la base du courant de la sociologie de la quantification (Bardet & Jany-Catrice, 2010; Martin, 2020) ont montré que les données (data) n'étaient pas…données : elles sont parfois difficiles à obtenir, elles ne sont pas directement interprétables, elles sont souvent transformées, « nettoyées » et mises en forme avant leur utilisation, au prix de modification parfois importantes (standardisation, « normalisation » par transformations mathématiques, élimination des valeurs extrêmes, etc.). Les données massives sont encore plus susceptibles d'être manipulées (les formations en science des données donnent aujourd'hui une large place à l'acquisition et au « nettoyage » des données non structurées).
À supposer même que les données disponibles soient traçables et que leur processus de transformation soit documenté, l'effet « boite noire » de certains algorithmes demeure un obstacle à la réplicabilité et à l'inférence logique. Certains algorithmes d'apprentissage supervisé et non supervisé sont en effet très sensibles aux paramétrages et aux conditions initiales : des résultats très différents peuvent être obtenus sur les mêmes commandes, tout particulièrement dans les méthodes de classification. Cela pose un problème de réplicabilité et « d'explicabilité » des résultats (Christin, 2017; Villani et al., 2018).

*2.4.3. Les risques d'un management « data driven »*
L'approche inductive inscrite dans les outils prédictifs d'analytique RH reposant sur des algorithmes est clairement orientée vers la prise de décision, avec un impact potentiel sur le sort des salariés ou candidats concernés. Un management et une prise de décision guidés par les données (*data driven management*) dans le domaine RH semblent au premier abord désirable en raison de « l'objectivité » et du caractère scientifique des méthodes employées. Celles-ci s'inscrivent dans le courant de *l'evidence based* management, en mettant au centre de l'argumentation la notion de preuve empirique (Coron, 2019, p. 43). Il nous faut toutefois demeurer prudents, car des travaux récents convergent vers la mise en évidence d'une série de risques et de dangers liés à ces pratiques, notamment ne matière de discrimination (O'Neil, 2017). Dans un article de synthèse, Giermindl et ses collègues dressent un panorama des menaces et du « côté obscur » du management par les données (Giermindl et al., 2021) : ils identifient six périls, parmi lesquels le réductionnisme (croyance naïve en la capacité des données massives à objectiver les phénomènes organisationnels), les prédiction auto-réalisatrices (lorsque les prédictions guident les décisions), le risque de dépendance de chemin (lorsque les prévisions faites sur le passé contraignent les décision pour le futur), ou encore la réduction de l'autonomie des salariés ou la marginalisation du raisonnement humain et des compétences managériales au profit des algorithmes (voir aussi Coron , 2019, p 119 pour une analyse des décisions fondées uniquement sur les chiffres).

## Conclusion

La quantification des RH est une pratique ancienne, avec des méthodes de reporting depuis longtemps institutionnalisées (Herrbach, 2021), mais force est de constater que le développement du management *data driven* apporte un renouvellement dans le domaine, en lien avec l'évolution des méthodes prédictives (apprentissage supervisé et non supervisé). Les pratiques des entreprises et des chercheurs en GRH sont concernées à parts égales par le phénomène, même si les premières se sont emparées du sujet plus rapidement que les seconds. Une phrase souvent reprise attribuée au mathématicien Clive Humby nous semble bien résumer les enjeux actuels : "Data is the new oil. It's valuable, but if unrefined it cannot really be used. It has to be changed into gas, plastic, chemicals, etc to create a valuable entity that drives profitable activity; so must data be broken down, analyzed for it to have value." (Mavuduru, 2022). Les données sont de plus en plus disponibles en volume et en variété, et le challenge à relever réside bien dans notre capacité à en extraire de la connaissance.

Les approches quantitatives inductives modernes peuvent permettre de transformer les données en connaissances valorisable : l'usage croissant du *data mining* dans les entreprises en apporte la preuve. Il pourrait en être de même dans la recherche académique, en visant la production de savoirs actionnables : l'histoire des sciences de l'organisation nous fournit de nombreux exemples de découvertes parfois fortuites fondées sur des données de terrain. On peut citer l'exemple de la découverte des facteurs d'hygiène et de motivation par F. Herzberg, qui s'est appuyé sur la méthode des incidents critiques et l'analyse factorielle pour faire émerger des facteurs distincts de satisfaction et d'insatisfaction (Herzberg et al., 2017, p. 13).

La capacité à produire cette connaissance à partir des données porte aussi des enjeux de légitimation de la fonction RH : le sujet des apports des RH à la performance des organisations structure toujours le champ du management stratégique des ressources humaines (Huselid, 1995; Wright et al., 2018), et l'analyse des facteurs humains prédictifs de la performance demeure un sujet central, dont les développements futurs ne pourront ignorer la révolution initiée par les big data (Ben-Gal, 2019).

**Annexe 1** : Les méthodes exploratoires (références et outils logiciels)

| MÉTHODES | OUTILS LOGICIELS |
|---|---|
| Corrélations | |
| Matrices de corrélations | R (R Core Team, 2021) <br> JASP (JASP Team, 2022) <br> JAMOVI (JAMOVI Project, 2021) |
| Corrélogrammes | Package psych (Revelle, 2021) |
| Corrélations polychoriques et tétrachoriques | Package psych |
| Tableaux croisés et coefficients d'association | R, JASP, JAMOVI |
| Régressions « exploratoires » et méthodes explicatives *data driven* | |
| Régressions hiérarchiques Ascendante/Descendante /Combinée (Stepwise) | Package gmulti (Calcagno & de Mazancourt, 2010) <br> JASP |
| Analyse de dominance | Package dominanceanalysis (Bustos Navarrete & Coutinho Soares, 2020) |
| Analyse d'importance relative | Package Relaimpo (Grömping, 2006) |
| Régressions pénalisées (Ridge, Lasso) | Package Ridge (Cule et al., 2022) <br> Package glmnet (Hastie et al., 2021) |
| Modèles prédictifs adaptatifs de classification (pour variable dépendante nominale) et de régression (pour variable dépendante continue) : Réseaux de neurones, forêts aléatoires,…. | Package caret (Kuhn et al., 2020) <br> JASP (module machine learning) |
| Méthode Fuzzy set QCA | Package QCA (Dușa, 2018) |
| Analyses factorielles | |
| Analyse en composantes principales (ACP) | Package FactoMiner (Husson, Josse, et al., 2016) <br> R/JASP / JAMOVI |
| Analyse factorielle des correspondances (AFC) | Package FactoMiner <br> R/JASP / JAMOVI |
| AFCM Analyse factorielle des correspondances multiples | Package FactoMiner <br> R/JASP / JAMOVI |
| Positionnement multidimensionnel (Multidimensional scaling) | Package MASS (Ripley & Venables, 2002) <br> R, JAMOVI |
| Analyses typologiques (clustering) | |
| Classifications hiérarchiques | Package cluster (Maechler et al., 2022) <br> JASP / JAMOVI |
| Centres mobiles | Package cluster |
| Modèles de mélange (mixture modeling) : classes latentes et profils latents | Package mclust (Scrucca et al., 2016) <br> Package tidyLPA (Rosenberg et al., 2019) |
| Analyse typologique longitudinales (analyses de séquences, analyse de trajectoires latentes) | Package TraMineR (Gabadinho et al., 2011) |
| Réseaux psychologiques | |
| Corrélations partielles | Package bootnet (Epskamp & Fried, 2020) <br> JASP (menu Network) |
| Modèles graphiques gaussiens (MGG) | Package bootnet |
| Modèle Ising | Package bootnet |
| Statistiques textuelles | |
| Lexicométrie | Package RTemis (Bouchet-Valat et al., 2021) <br> Package Quanteda (Benoit et al., 2018) |
| Classification descendante (méthode Reinert) | Package.Rainette (Barnier, 2022) <br> Iramuteq (Ratinaud, 2020) |
| Analyse de sentiment | Package Quanteda <br> Package SentimentAnalysis (Proellochs, 2021) |
| Segmentation prédictive | |
| *Predicted oriented segmentation* | Package plspm (Sanchez, 2013) |

**Annexe 2** : Validation du modèle de mesure

**Mesure de l'implication unidimensionnelle et multi-cible** (Klein & al, 2014). Réponses sur une échelle à 5 degrés (Pas du tout, Peu, Modérément, Plutôt, Extrêmement)

**Cible organisation (KUTO)**

1. Jusqu'à quel point êtes-vous impliqué(e) envers votre organisation ?
2. Dans quelle mesure vous préoccupez-vous de votre organisation ?
3. Jusqu'à quel point êtes-vous dévoué(e) envers votre organisation ?
4. Dans quelle mesure avez-vous choisi d'être impliqué(e) envers votre organisation ?

**Cible Collègues (KUTC)**

1. Jusqu'à quel point êtes-vous impliqué(e) envers vos collègues ?
2. Dans quelle mesure vous préoccupez-vous de vos collègues ?
3. Jusqu'à quel point êtes-vous dévoué(e) envers vos collègues ?
4. Dans quelle mesure avez-vous choisi d'être impliqué(e) envers vos collègues ?

**Cible Profession (KUTP)**

1. Jusqu'à quel point êtes-vous impliqué(e) envers votre métier/profession ?
2. Dans quelle mesure vous préoccupez-vous de votre métier/profession ?
3. Jusqu'à quel point êtes-vous dévoué(e) envers votre métier/ profession ?
4. Dans quelle mesure avez-vous choisi d'être impliqué(e) envers votre métier/profession ?

**Tableau .** Analyse confirmatoire des échelles de mesure

| Échelle | Alpha | Contributions factorielles (*loadings*) | Variance moy extraite (AVE) |
|---|---|---|---|
| KUTO (impl. organisationnelle) 4 items | 0.89 | min : 0.81 ; max 0.84 | 0.54 |
| KUTC (impl. vers les collègues) 4 items | 0.90 | min : 0.81 ; max 0.85 | 0.50 |
| KUTPR (impl. professionnelle) 4 items | 0.91 | min : 0.82 ; max 0.88 | 0.53 |
| satisfaction au travail 3 items | 0.82 | min : 0.71 ; max 0.81 | 0.60 |
| Intention de quitter 4 items | 0.88 | min : 0.77 ; max 0.91 | 0.65 |
| Identification organisationnelle 6 items | 0.88 | min : 0.65 ; max 0.82 | 0.55 |
| Engagement dans le travail 9 items | 0.93 | min : 0.69 ; max 0.87 | 0.61 |

Indicateurs d'ajustement du modèle global (Analyse factorielle confirmatoire, estimateur MLR, adapté aux variables non -normales) : Khi² (390) = 927, *p* <.001 ; CFI = 0.96 ; TLI = 0.95 ; RMSEA = 0.05 [0.042, 0.052] ; SRMR = 0.03

**Annexe 3 :** Analyse en profils latents

Une fois déterminées les variables de classification, on doit sélectionner un modèle d'estimation et un nombre de classes. Les modèles d'estimation diffèrent selon les *contraintes imposées à la distribution des données dans les profils*. Une procédure itérative basée sur le test de différentes solutions avec calcul d'indices d'ajustement permet de guider vers le choix d'un nombre de classes optimal.

Dans une logique exploratoire, si on sélectionne trop peu de classes, la solution sera peu intéressante et si on sélectionne trop de classes, elle sera non interprétable. Nous choisissons donc une plage de test entre 3 et 6 classes, avec à chaque fois le test des 6 modèles de mélange disponibles dans le package tidy LPA. Ces modèles diffèrent dans les contraintes apportées sur la variance des profils et la covariance entre les profils (variances contraintes à l'égalité ou librement estimées, covariances contraintes à zéro, contraintes à l'égalité ou librement estimées). Les modèles les plus contraints sont aussi les plus parcimonieux (Bauer, 2021).

Une analyse hiérarchique sur une série d'indicateurs selon la procédure AHP proposée par Akogul & Erisoglu (2017) implémentée dans le package tidyLPA (fonction *compare_solutions*) conduit à recommander le modèle 4 avec 5 classes. Il s'agit d'un modèle avec variances libres et covariance fixées à zéro.

| Modèle | Nb de Classes | LogLik | AIC | BIC | KIC | SABIC | ICL | Entropie |
|---|---|---|---|---|---|---|---|---|
| 1 | 3 | -3302,50 | 6632,99 | 6699,42 | 6649,99 | 6654,96 | -6899,67 | 0,79 |
|   | 4 | -3227,61 | 6491,21 | 6576,63 | 6512,21 | 6519,47 | -6717,43 | 0,88 |
|   | 5 | -3198,61 | 6441,22 | 6545,61 | 6466,22 | 6475,75 | -6694,52 | 0,89 |
|   | 6 | -3157,41 | 6366,82 | 6490,19 | 6395,82 | 6407,62 | -6602,81 | 0,92 |
| 2 | 3 | -2964,74 | 5969,48 | 6064,39 | 5992,48 | 6000,87 | -6225,63 | 0,82 |
|   | 4 | -2783,27 | 5620,54 | 5748,66 | 5650,54 | 5662,92 | -5897,36 | 0,86 |
|   | 5 | -2669,60 | 5407,19 | 5568,53 | 5444,19 | 5460,55 | -5715,76 | 0,88 |
| 3 | 3 | -3219,85 | 6473,70 | 6554,37 | 6493,70 | 6500,39 | -6607,31 | 0,93 |
|   | 4 | -3198,44 | 6438,89 | 6538,54 | 6462,89 | 6471,85 | -6662,45 | 0,88 |
|   | 5 | -3189,98 | 6429,95 | 6548,58 | 6457,95 | 6469,19 | -6685,03 | 0,88 |
|   | 6 | -3178,42 | 6414,84 | 6552,45 | 6446,84 | 6460,36 | -6968,24 | 0,75 |
| 4 | 4 | -2753,05 | 5566,11 | 5708,46 | 5599,11 | 5613,19 | -5864,81 | 0,86 |
| **4** | **5** | **-2397,38** | **4868,76** | **5044,34** | **4908,76** | **4926,84** | **-5218,98** | **0,86** |
| 6 | 3 | -2974,68 | 6007,36 | 6144,97 | 6039,36 | 6052,87 | -6347,05 | 0,74 |
| 6 | 4 | -2599,14 | 5276,27 | 5461,34 | 5318,27 | 5337,49 | -5640,35 | 0,83 |
| 6 | 5 | -2487,16 | 5072,32 | 5304,83 | 5124,32 | 5149,22 | -5563,46 | 0,81 |

*Interprétation des indices* : pour tous les indices à l'exception de l'entropie, une valeur plus basse indique un meilleur ajustement. L'entropie peut s'interpréter comme un analogue d'un coefficient de fiabilité (Alpha de Cronbach) : une valeur > 0.8 indique un bon ajustement du modèle
*nb. Certaines estimations n'ont pas convergé, ce qui explique que le tableau ne contienne pas l'ensemble des solutions possibles (24)*

On vérifie ensuite que la solution est suffisamment équilibrée en affichant le nombre d'observations par classes. C'est le cas ici, avec des effectifs respectifs de 159, 84, 179, 281, et 147 individus

**Annexe 4 :** Exploration des profils

Le package FactoMiner fournit une fonction (catdes) qui permet de comparer rapidement et efficacement les classes en utilisant des variables illustratives qualitatives et quantitatives. Un test de significativité (v.test) fonctionne comme une analyse de contraste, et permet de mettre en évidence les différences entre profils pour les variables illustratives. Pour les variables quantitatives, le score moyen sur le profil est est comparé à la moyenne générale, et pour les variables qualitative, la répartiiton des modalités est comparée à une répartition au hasard.

**Tableau.** Description des 5 profils par les variables illustratives

| Profil1 | | | | | | | | | |
|---|---|---|---|---|---|---|---|---|---|
| **Variables Quantitatives** | v.test | *p* | Moy dans le profil | Moyenne générale | **Variables Qualitatives** | v.test | *p* | % dans le profil | % dans l'ensemble |
| Turnover | 6,59 | *** | 0,50 | 0,00 | Ancienn. : <1 an | 4,01 | *** | 20,13 | 10,59 |
| AGE | -2,55 | *** | 38,88 | 40,87 | Pays : Canada | 2,75 | ** | 34,59 | 25,76 |
| Identification | -10,15 | *** | -0,77 | 0,00 | Ancienn. : 5 à 10 | -2,27 | * | 16,35 | 23,06 |
| Satisfaction | -11,01 | *** | -0,86 | 0,00 | Pays : France | -2,81 | *** | 16,98 | 25,53 |
| Engagement | -11,30 | *** | -0,83 | 0,00 | | | | | |

| Profil 2 | | | | | | | | | |
|---|---|---|---|---|---|---|---|---|---|
| Turnover | 2,11 | ** | 0,23 | 0,00 | Pays : Canada | 2,62 | ** | 38,10 | 25,76 |
| Identification | -4,27 | *** | -0,47 | 0,00 | Pays : Suisse | 2,04 | * | 34,52 | 25,06 |
| Satisfaction | -4,87 | *** | -0,55 | 0,00 | Pays : France | -4,13 | *** | 8,33 | 25,53 |
| Engagement | -6,19 | *** | -0,66 | 0,00 | | | | | |

| Profil 3 | | | | | | | | | |
|---|---|---|---|---|---|---|---|---|---|
| AGE | 2,19 | * | 42,46 | 40,87 | | | | | |
| Satisfaction | 2,08 | * | 0,15 | 0,00 | | | | | |

| Profil 4 | | | | | | | | | |
|---|---|---|---|---|---|---|---|---|---|
| Identification | 2,48 | ** | 0,13 | 0,00 | Pays : France | 3,18 | *** | 32,38 | 25,53 |
| Satisfaction | 2,35 | * | 0,12 | 0,00 | Pos. : Manager | 3,12 | *** | 27,76 | 21,41 |
| Engagement | 2,26 | * | 0,11 | 0,00 | | | | | |

| Profil 5 | | | | | | | | | |
|---|---|---|---|---|---|---|---|---|---|
| Engagement | 12,72 | *** | 0,98 | 0,00 | Pays : France | 2,13 | * | 32,65 | 25,53 |
| Satisfaction | 10,03 | *** | 0,82 | 0,00 | Ancienn. : < 1 an | -2,00 | * | 6,12 | 10,59 |
| Identification | 9,53 | *** | 0,76 | 0,00 | Pays : Canada | -2,30 | * | 18,37 | 25,76 |
| Turnover | -8,63 | *** | -0,68 | 0,00 | Taille : > 1000 | -2,82 | *** | 12,93 | 21,29 |

*Lecture du tableau : le zones colorées signalent une moyenne significativement plus faible pour les variables quantitatives ou une sous-représentation pour les variables qualitatives*

**Annexe 5 :** Corrélations entre les variables continues du jeu de données et les probabilités d'appartenance aux classes

| V1 | | V2 | r | |
|---|---|---|---|---|
| Probabilité d'appartenance au profil 1 | - | Implication orga. | -0,55 | *** |
| | - | Implication collègues | -0,46 | *** |
| | - | Implication profession | -0,65 | *** |
| | - | Satisfaction | -0,43 | *** |
| | - | Turnover | 0,26 | *** |
| | - | Identification | -0,41 | *** |
| | - | Engagement | -0,46 | *** |
| Probabilité d'appartenance au profil 2 | - | Implication orga. | -0,36 | *** |
| | - | Implication collègues | -0,26 | *** |
| | - | Implication profession | -0,30 | *** |
| | - | Satisfaction | -0,17 | *** |
| | - | Turnover | 0,08 | * |
| | - | Identification | -0,15 | *** |
| | - | Engagement | -0,22 | *** |
| Probabilité d'appartenance au profil 3 | - | Implication orga. | 0,03 | |
| | - | Implication collègues | 0,04 | |
| | - | Implication profession | 0,06 | |
| | - | Satisfaction | 0,07 | |
| | - | Turnover | -0,03 | |
| | - | Identification | 0,04 | |
| | - | Engagement | 0,03 | |
| Probabilité d'appartenance au profil 4 | - | Implication orga. | 0,12 | *** |
| | - | Implication collègues | 0,25 | *** |
| | - | Implication profession | 0,28 | *** |
| | - | Satisfaction | 0,10 | ** |
| | - | Turnover | 0,03 | |
| | - | Identification | 0,11 | ** |
| | - | Engagement | 0,10 | ** |
| Probabilité d'appartenance au profil 5 | - | Implication orga. | 0,62 | *** |
| | - | Implication collègues | 0,32 | *** |
| | - | Implication profession | 0,46 | *** |
| | - | Satisfaction | 0,35 | *** |
| | - | Turnover | -0,30 | *** |
| | - | Identification | 0,33 | *** |
| | - | Engagement | 0,44 | *** |

**Annexe 6** : Modèles graphiques gaussiens

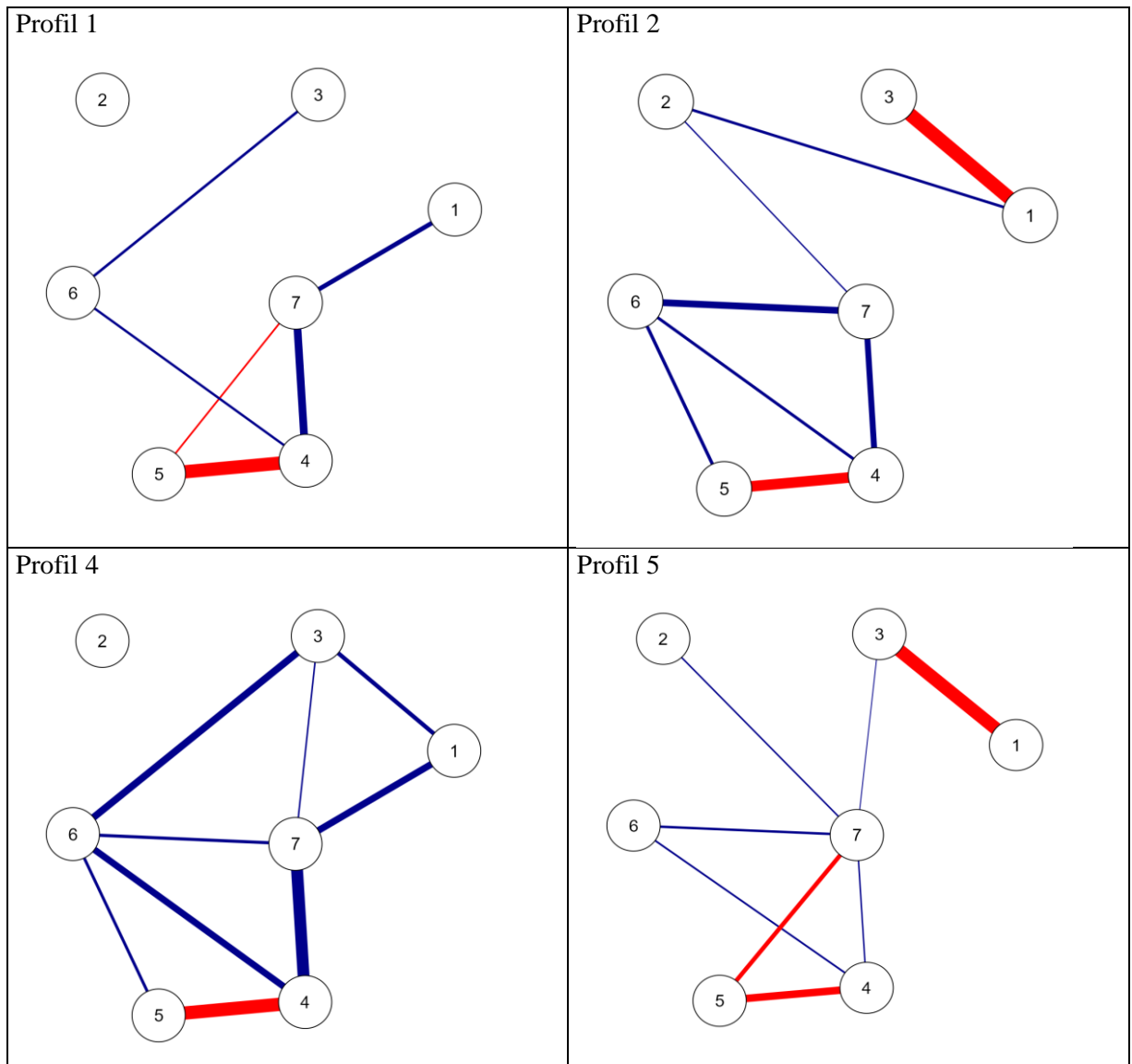

Légende

1: impli orga.
2: impli collegues
3: impli profession
4: Satisfaction
5: Turnover
6: Identification
7: Engagement

*Lecture :* les traits matérialisent des corrélations partielles. l'épaisseur des traits signale la magnitude de la corrélation et la couleur le sens de la corrélation (rouge pour les corrélations négatives et bleu pour les corrélations positives).

**Indicateurs de centralité**

Les indicateurs de centralité sont des indices permettant l'évaluer l'importance rôle joué par une variable dans un réseau. Il en existe plusieurs, dont les plus utilisés sont le degré de centralité et la centralité d'intermédiation. Pour plus de détails, on peut se reporter à (Costantini et al., 2015; Dalege et al., 2017).

| Indicateur | Définition / Interprétation / Exemple |
|---|---|
| Degré de centralité (*Degree / Strength*) | Définition : Dans un MGG, le degré de centralité (*Strength*) dite aussi centralité de degré correspond à la somme des poids des liens connectés à un nœud.<br>Interprétation : une forte centralité signale une variable jouant un rôle important dans le réseau, grâce à des liens directs forts (sans médiation) avec d'autres variables.<br>*Exemple : sur le profil n°1 le nœud n°4 (satisfaction) a le plus fort degré de centralité du réseau car la variable est fortement liée à ses voisins* |
| Centralité d'intermédiation (*Betweenness centrality*) | Définition : Nombre de plus courts chemins (*shortest paths*) transitant par un nœud.<br>Interprétation : une variable avec une forte centralité d'intermédiation joue un rôle de relais dans la transmission de l'information entre différentes parties du réseau (variable médiatrice).<br>*Exemple : sur le profil 5, le nœud n°7 (engagement) a le plus fort coefficient de centralité d'intermédiation car se trouve sur le passage de nombreux chemins.* |

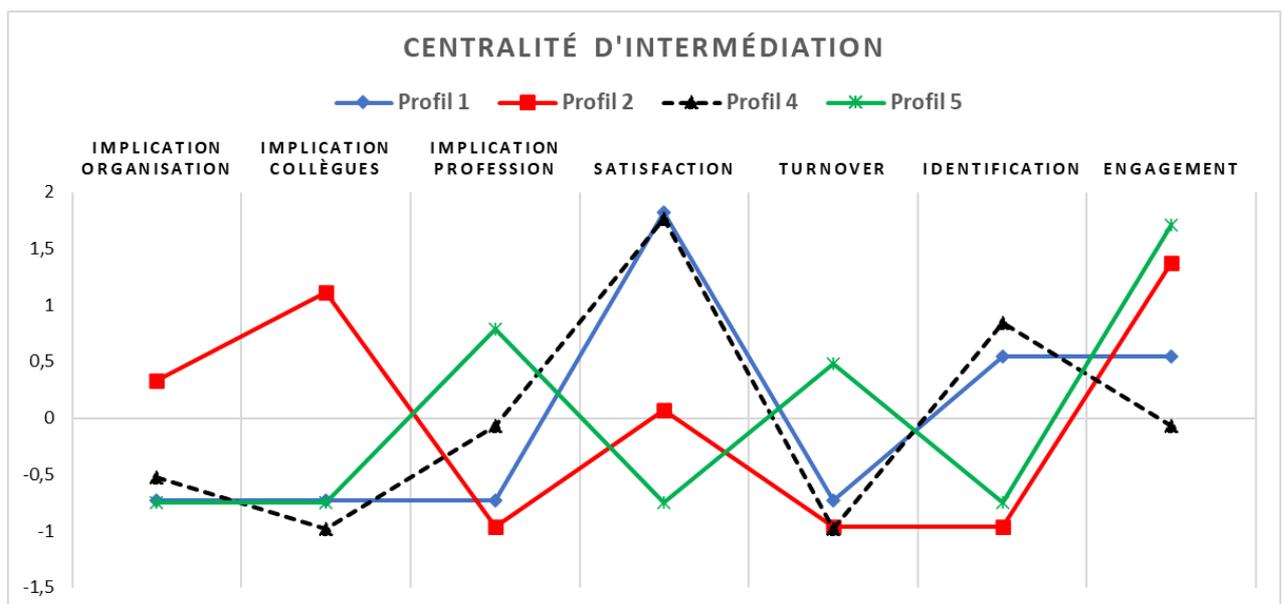

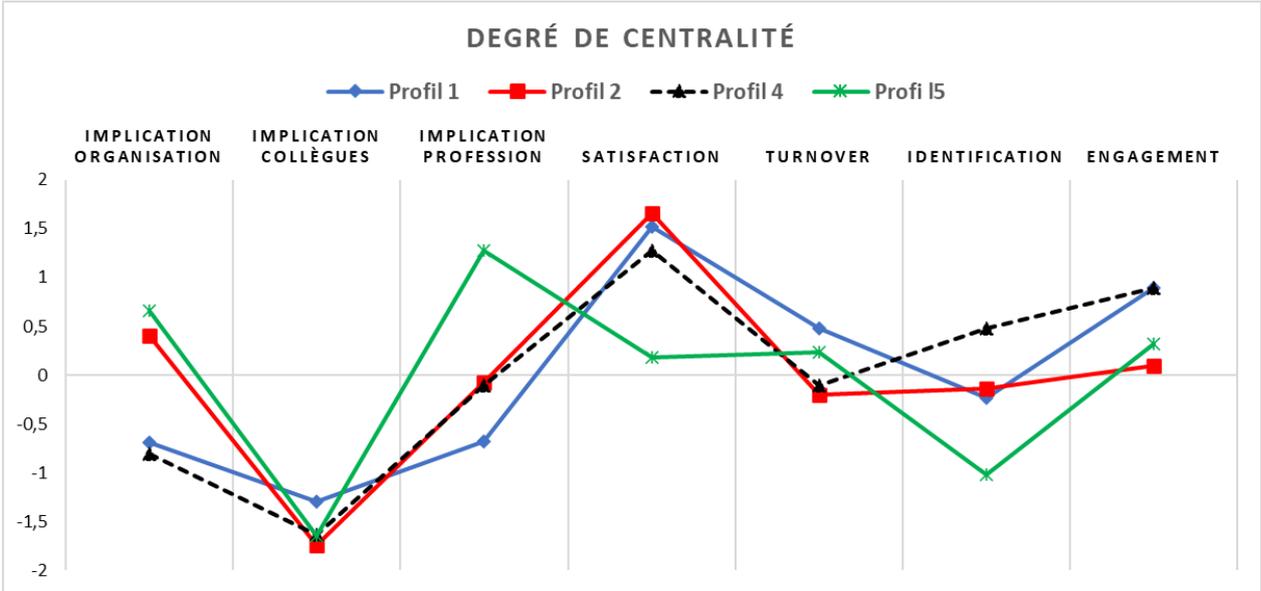

**Annexe 7 :** Synthèse

| Segmentation | L'analyse en profils latents suggère une répartition en 5 profils d'implication. Les profils sont organisés selon le niveau d'implication dans les trois cibles : cela suggère l'existence d'un construit d'implication globale d'ordre supérieur. Ce construit pourrait être mis en évidence par l'estimation d'un modèle bi-factoriel.<br>Un examen des prédicteurs amène à constater que le profil médian (profil 3) est peut-être artificiel, car la capacité des variables du jeu de données à prédire l'appartenance à ce profil est très faible ($R^2 = 0.02$). |
|---|---|
| Profil 1 | « Les insatisfaits / non impliqués »<br>Répondants peu impliqués dans les trois cibles, avec une grande distance vis à vis de l'organisation et une forte intention de quitter<br>Sur-représentation des répondants canadiens.<br>Prédicteur principal : la faible implication dans la profession (toute hausse de l'implication dans la profession réduit la probabilité d'appartenir au profil 1).<br>La satisfaction (ici l'insatisfaction) joue un rôle de passerelle important dans le réseau relationnel. |
| Profil 2 | « Les détachés / résignés »<br>Profil très proche du profil 1 sur le niveau d'implication dans les trois cibles, avec une intention de quitter plus faible que les salariés du profil 1.<br>Prédicteur principal : la faible implication dans l'organisation (toute hausse de l'implication dans l'organisation réduit la probabilité d'appartenir au profil 2). On peut envisager ici des situations où le travail est avant tout utilitaire.<br>Les cibles organisation et profession semblent substituables (corrélation négative forte), ce qui suggère deux sous-profils : les peu impliqués avant tout dans leur métier ou avant tout dans l'organisation et les collègues (lien positif entre les 2 cibles). |
| Profil 4 | « Les bons collègues »<br>Les répondants appartenant à ce profil présentent un profil d'implication moyen dans les trois cibles avec une satisfaction et un engagement qui sont supérieurs à la moyenne.<br>Prédicteur principal : L'implication dans la profession (toute hausse de l'implication dans la profession augmente la probabilité d'appartenir au profil 4). L'implication envers les collègues est aussi un prédicteur important dans ce profil<br>Les cibles organisation et profession sont positivement liées, ce qui suggère une implication générale (métier et emploi occupé ne sont pas incompatibles).<br>La satisfaction joue un rôle de passerelle important dans le réseau relationnel |
| Profil 5 | « Les impliqués / engagés »<br>Les répondants appartenant à ce profil sont les plus impliqués dans les trois cibles et ont une forte intention de rester. Les managers et les répondants française sont surreprésentés dans ce profil.<br>Prédicteurs principaux : L'implication dans l'organisation et l'engagement au travail<br>Les cibles organisation et profession semblent substituables (corrélation négative forte), ce qui suggère deux sous-profils : les impliqués avant tout dans leur métier ou avant tout dans l'organisation. L'implication dans la profession et l'organisation ainsi que l'engagement occupent une place centrale dans le réseau relationnel. |